\pdfoutput=1
\documentclass[11pt]{article}

\usepackage{emnlp2021}
\usepackage{times}
\usepackage{latexsym}

\usepackage[T1]{fontenc}
\usepackage[utf8]{inputenc}

\usepackage{microtype}

\usepackage{times}
\usepackage{latexsym}
\usepackage{latexsym}
\usepackage{booktabs}
\usepackage{amsmath}

\usepackage{amssymb}
\usepackage{mathtools}
\usepackage{array}
\usepackage{multirow}
\usepackage{lipsum}
\usepackage{bm}
\usepackage{subfigure}
\usepackage{algorithmicx}
\usepackage{algpseudocode}
\usepackage{microtype}
\usepackage{graphicx}
\usepackage[ruled,vlined]{algorithm2e}

\usepackage{microtype}

\title{
CascadeBERT: Accelerating Inference of Pre-trained Language Models via Calibrated Complete Models Cascade
}

\author{Lei Li\textsuperscript{$\dag$}, Yankai Lin\textsuperscript{$\S$}, Deli Chen\textsuperscript{$\dag\S$}, Shuhuai Ren\textsuperscript{$\dag$}, Peng Li\textsuperscript{$\S$}, Jie Zhou\textsuperscript{$\S$}, Xu Sun\textsuperscript{$\dag$}\\
   \textsuperscript{$\dag$}MOE Key Laboratory of Computational Linguistics, School of EECS, Peking University \\
  \textsuperscript{$\S$}Pattern Recognition Center, WeChat AI, Tencent Inc., China\\
    \texttt{\{lilei, shuhuai\_ren\}@stu.pku.edu.cn} \\
    \texttt{\{chendeli, xusun\}@pku.edu.cn} \\
    \texttt{\{yankailin, patrickpli, withtomzhou\}@tecent.com}
  }
\begin{document}
\maketitle
\begin{abstract}
Dynamic early exiting aims to accelerate the inference of pre-trained language models (PLMs) by emitting predictions in internal layers without passing through the entire model.
In this paper, we empirically analyze the working mechanism of dynamic early exiting and find that it faces a performance bottleneck under high speed-up ratios. 
On one hand, the PLMs' representations in shallow layers lack high-level semantic information and thus are not sufficient for accurate predictions. 
On the other hand, the exiting decisions made by internal classifiers are unreliable, leading to wrongly emitted early predictions.
We instead propose a new framework for accelerating the inference of PLMs, CascadeBERT, which dynamically selects proper-sized and complete models in a cascading manner, providing comprehensive representations for predictions. 
% To tackle these problems, we propose CascadeBERT, which dynamically selects proper-sized and complete models in a cascading manner, providing comprehensive representations for predictions. 
We further devise a difficulty-aware objective, encouraging the model to output the class probability that reflects the real difficulty of each instance for a more reliable cascading mechanism.
Experimental results show that CascadeBERT can achieve an overall 15\% improvement under 4$\times$ speed-up compared with existing dynamic early exiting methods on six classification tasks, yielding more calibrated and accurate predictions.\footnote{Our code is available at \url{https://github.com/lancopku/CascadeBERT}}

% Further analysis show that the difficulty-aware regularization helps calibrate model predictions.
\end{abstract}
\section{Introduction}
\label{sec:intro}
Large-scale pre-trained language models~(PLMs), e.g., BERT and RoBERTa, have demonstrated superior performance on various natural language understanding tasks~\citep{devlin2019bert,Liu2019RoBERTa}.
% , including sentiment analysis~\cite{sun2019utilizing}, question answering.
While the increased model size brings more promising results, the long inference time hinders the deployment of PLMs in real-time applications.
Researchers have recently exploited various kinds of approaches for accelerating the inference of PLMs, which can be categorized into model-level compression and instance-level speed-up.
The former aims at obtaining a compact model via quantization~\citep{zafrir2019q8bert,Shen2020QBERT,zhang2020ternarybert},  pruning~\citep{voita2019analyzing,michel2019sixteen} or knowledge distillation~(KD)~\citep{Sanh2019DistilBERT,Sun2019PatientKD,Jiao2019TinyBERT}, while the latter adapts the amount of computation to the complexity of each instance~\citep{graves2016adaptive}.
% treats each instance differently based on the .
% 目前主流的方法就是 dynamic early exiting
A mainstream method for instance-level speed-up is dynamic early exiting, which emits predictions based on intermediate classifiers~(or off-ramps) of internal layers when the predictions are confident enough~\citep{Xin2020DeeBERT,Liu2020FastBERT,schwartz-etal-2020-right,li2021accelerating}. 
\begin{figure}[t!]
    \centering
    \includegraphics[width=0.99\linewidth]{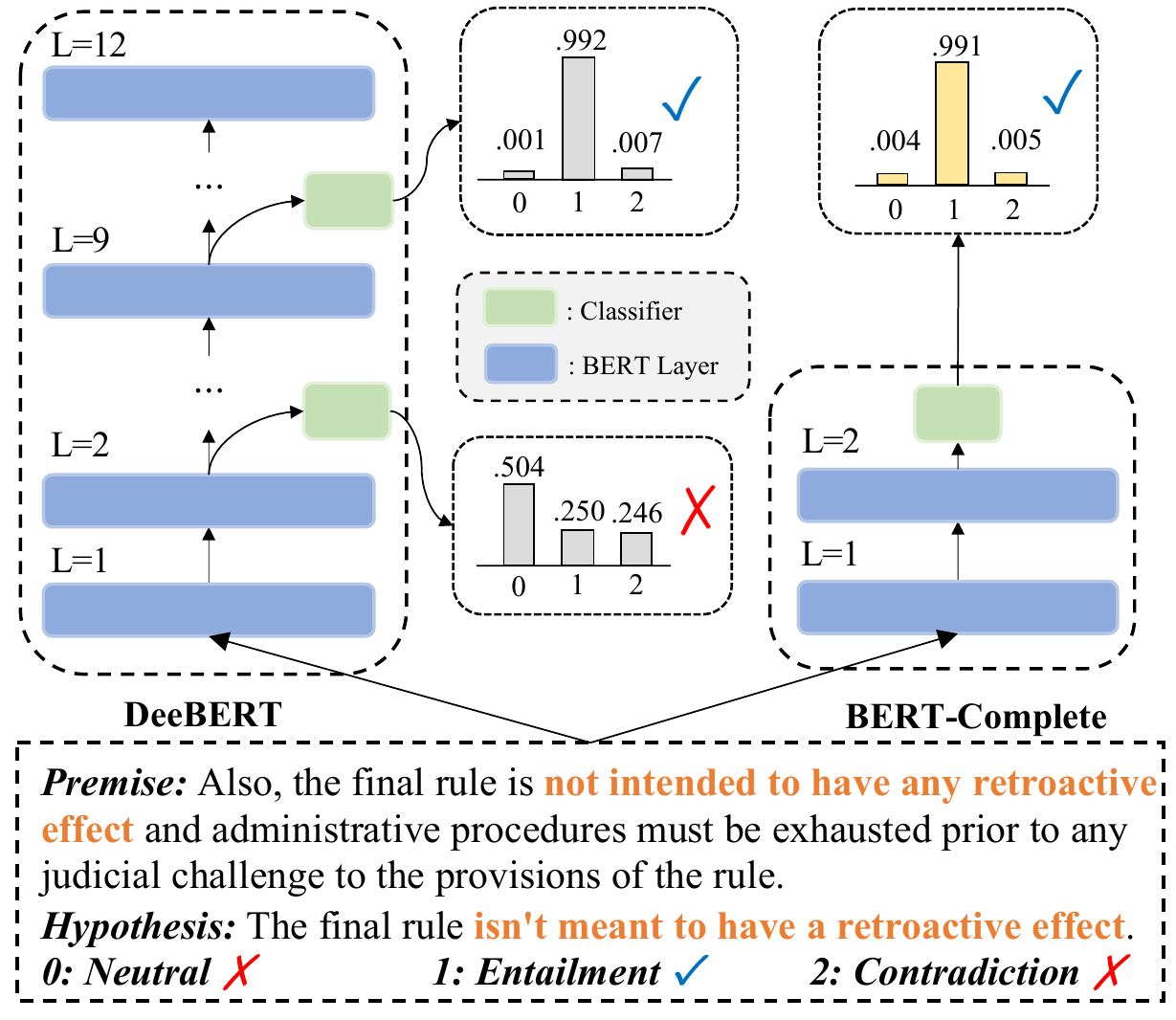}
    \caption{An easy instance with a large word overlap (colored in orange) between the premise and the hypothesis from the MNLI dataset. The classifiers in shallow layers of a dynamic early exiting model cannot predict correctly, while BERT-Complete~\citep{turc2019well}, a small BERT pre-trained from scratch with the same size can make a correct and confident prediction.}
    \label{fig:easy_instance}
\end{figure}
% 

% Although the idea is intuitive and simple, dynamic early exiting 
In this paper, we focus on dynamic early exiting, as it can be utilized to accelerate inference and reduce the potential risk of the overthinking problem~\citep{kaya2019overthinking}. 
Such a paradigm is intuitive and simple, while faces a performance bottleneck under high speed-up ratios, i.e., the task performance is poor when most examples are exited in early layers. 
We conduct probing experiments to investigate the mechanism of dynamic exiting, and find that the poor performance is due to the following two reasons: 
(1) The shallow representations lack high-level semantic information, and are thus not sufficient for accurate predictions.
As PLMs like BERT exhibit a hierarchy of representations, e.g., shallow layers extract low-level features like lexical/syntactic information while deep layers capture semantic-level relations~\citep{tenney-etal-2019-bert,jawahar2019structure}, we argue that the high-level semantic inference ability is usually required even for easy instances.
As shown in Figure~\ref{fig:easy_instance}, the classifier of the second layer in a representative early exiting model DeeBERT~\cite{Xin2020DeeBERT} cannot predict correctly even for an easy instance with a large word overlap.
On the contrary, BERT-Complete, a shallow 2-layer model pre-trained from scratch~\citep{turc2019well} that is thus capable of extracting semantic-level features, can make confident and correct predictions like that in deep layers of DeeBERT. 
(2) The intermediate classifiers in the early exiting models cannot provide reliable exiting decisions.
We design a metric to examine the ability of models to distinguish difficult instances from easy ones, which can reflect the quality of exiting decisions.
We find that the predictions of internal classifiers cannot faithfully reflect the instance difficulty, resulting in wrongly emitted results and thus hindering the efficiency of early exiting.
% to measure the quality of exiting decisions, we design a metric for examining model's ability to distinguish difficult instances from easy ones. 
% For example, a intermediate classifier at the second layer may produce wrong prediction with a nearly 100\% confidence score.
% \citet{desai2020calibration} also demonstrate that pre-trained language models also need calibration.

%\vskip -2pt
% To remedy those drawbacks, we propose \textbf{CascadeBERT},
% which conducts inference based
% on a series of complete models in a cascading manner with a dynamic stopping mechanism. 
To remedy those drawbacks, we instead extend the dynamic early exiting idea to a model cascade, and propose \textbf{CascadeBERT},
which conducts inference based
on a series of complete models in a cascading manner with a dynamic stopping mechanism. 
Specifically, given an instance for inference, instead of directly exiting in the middle layers of a single model, the framework progressively checks if the instance can be solved by the current PLM from the smallest to the largest one, and emits the prediction once the PLM is confident about the prediction. 
Furthermore, we propose a difficulty-aware regularization to calibrate the PLMs' predictions according to the instance difficulty, making them reflect the real difficulty of each instance. Therefore, the predictions can be utilized as a good indicator for the early stopping in inference.
Experimental results on six classification tasks in the GLUE benchmark demonstrate that our model can obtain a much better task performance than previous dynamic early exiting baselines under high speed-up ratios.
% (15.5\% relative improvement under 4$\times$ speed-up).
Further analysis demonstrates that the proposed difficulty-aware objective can calibrate the model predictions, and proves the effectiveness and the generalizability of CascadeBERT.

% In summary, our contributions are three-fold:
% \begin{itemize}
%     \item We examine the prevailing dynamic exiting framework for pre-trained language models acceleration. We find that shallow representations are not sufficient for accurate predictions and early classifiers cannot provide reliable exiting decisions, resulting in poor performance under high speed-up ratio.
%     \item We propose CascadeBERT, a framework to progressively conduct inference based on a series of complete models, and a difficulty-aware regularization objective for enhancing models' ability to make more robust emitting decisions.
%     \item Experimental results demonstrate the effectiveness of the proposed fraemwork, achieving 15\% relative improvement over the na\"ive early exiting method and obtaining superior results over knowledge distillation methods.
% \end{itemize}

\section{Investigations into Early Exiting}
% 介绍 Early exit 的框架 
Dynamic early exiting aims to speed-up the inference of PLMs by emitting predictions based on internal classifiers.
% by adding internal classifiers after each layer in the original model.
For each instance, if the internal classifier's prediction based on the current layer representation of the instance is confident enough, e.g., the maximum class probability exceeds a threshold~\citep{schwartz-etal-2020-right}, then the prediction is emitted without passing through the entire model. 
However, whether the internal representations could provide sufficient information for accurate predictions and whether the intermediate classifiers can be utilized for making accurate exiting decisions still remain unclear. 
In this section, we investigate the working mechanism of dynamic early exiting by exploring these two questions.
% In this section, we conduct probing experiments to examine whether the internal representations can provide a sufficient information for high-performance predictions and whether intermediate classifiers can be utilized for making exiting decisions.
\label{sec:analysis}
\subsection{Are Shallow Features Sufficient?}
\label{subsec:rep}
As discussed by \citet{tenney-etal-2019-bert}, PLMs like BERT learn a hierarchy of representations.
% and rediscover the traditional text processing pipeline.
% , e.g., basic syntactic information emerges in shallow layers, while deeper layers mainly capture high-level semantic structures. 
We assume that the high-level semantics is usually required even for easy instances, and therefore the predictions based on shallow representations are insufficient for accurate predictions. 
To examine this, 
% whether the shallow representations are sufficient, i.e., without the high-level semantic information (or the complete pipeline ability), for making accurate predictions,
we evaluate the model performance based on outputs of different layers,
% 缺了一句我们可以通过 分类器的性能来判断 表示是否充足
as the representation contains adequate information is necessary for a decent task performance.
% If the representation from a shallow layer contains adequate information for a task, it is reasonable to expect a classifier can achieve a decent performance based on it.
Specifically, we compare the following models:
%They also show that high-level information can be utilized to disambiguate low-level decisions like part-of-speech tagging.
% Their findings in all show that a pipeline system is recovered by the pre-trained language models like BERT.

%\medskip
\noindent\textbf{DeeBERT}~\citep{Xin2020DeeBERT}, which is a representative of early exiting methods. The internal classifiers are appended after each layer in the original BERT for emitting early predictions.
% Note that we emit \emph{all} instances with internal classifiers at different layers in DeeBERT, to examine the quality of predictions based on shallow representations.

%\medskip
\noindent\textbf{BERT-\textit{k}L}, which only utilizes the first $k$ layers in the original BERT model for prediction. A classifier is added directly after the first $k$ layers.
The parameters of the first $k$ layers and the classifier are fine-tuned on the training dataset. It could be seen as a static early exiting method.
% By comparing the performance of this method to DeeBERT, we can ablate the information gain in the fine-tuning process.

%\medskip
\noindent\textbf{BERT-Complete}~\citep{turc2019well}, which is a light version of the original BERT model pre-trained from scratch using the masked language modeling~(MLM) objective. 
We assume the representations of this model contain high-level semantic information, as MLM requires a deep understanding of the language.
% complete text processing pipeline ability.
% to explore the model performance with the full text processing pipeline.
% To simulate the full pipeline with  fewer model layers, we utilize the small versions of BERT provided by~\citet{turc2019well}. 
\begin{figure}[t]
    \centering
    \includegraphics[width=0.98\linewidth]{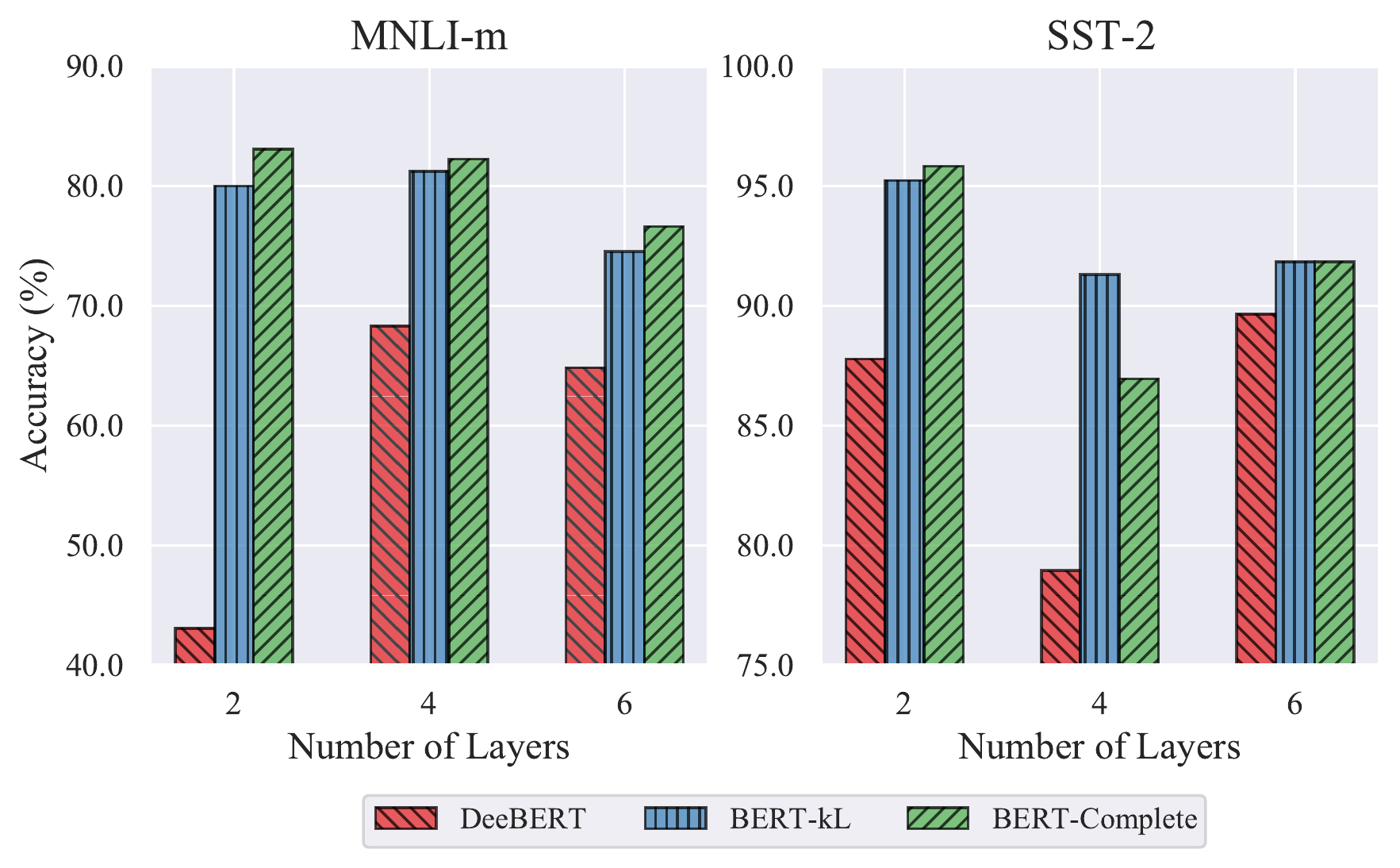}
    \caption{Performance comparison utilizing different models with the same number of layers on MNLI-m and SST-2. Complete models capable of extracting semantic-level information clearly outperform models like DeeBERT which overlooks the high-level semantic features.}
    \label{fig:pipeline}
\end{figure}

For a fair comparison, models are evaluated on a subset of instances which DeeBERT chooses to emit at different layers.
We report prediction accuracy using different number of layers on MNLI~\citep{williams2018mnli} and SST-2~\citep{socher2013sst}. 
% The former is a natural language inference task requiring the model to predict the relation between a premise and a hypothesis, while the latter is a dataset for sentiment classification.
% We evaluate the performance of using different number of layers on the MNLI dataset~\citep{williams2018mnli}, a natural language inference task requiring the model to predict the relation between a premise sentence and a hypothesis sentence.
Figure~\ref{fig:pipeline} shows the results on the development sets, and we can see that: 

% 首先说明 Pipeline 的重要性 gap 很大
 (1) BERT-Complete clearly outperforms DeeBERT, especially when the predictions are made based on shallow layers. 
 It indicates that the high-level semantics is vital for handling tasks like sentence-level classification.
% BERT-NL 效果也很不错 来自于 fine-tuning 对最后一层的改变

(2) BERT-\textit{k}L also outperforms DeeBERT. We attribute it to that the last serveral layers can learn task-specific information during fine-tuning to obtain a decent performance. A similar phenomenon is also observed by~\citet{Merchant2020ftEmbedding}. 
However, since the internal layer representation in DeeBERT are restricted by the layer relative position in the whole model, this adaption effect cannot be fully exploited, resulting in the poor performance in shallow layers.

These findings verify our assumption that the semantic-level features are vital, motivating us to exploit complete models for predictions.
Besides, DeeBERT performs poorly on the selected instances which it decides to emit at different layers, triggering our further explorations on the quality of exiting decisions.
% Gap 逐渐缩小，说明高层的信息是有用的
% (3) The gap is narrowed as the number of layer increases,
% which in turn validates our assumption that shallow representations are not sufficient for accurate predictions.
\subsection{Are Internal Classifiers Reliable?}
\label{subsec:dis}
\begin{figure}[t!]
    \centering
    \includegraphics[width=0.98\linewidth]{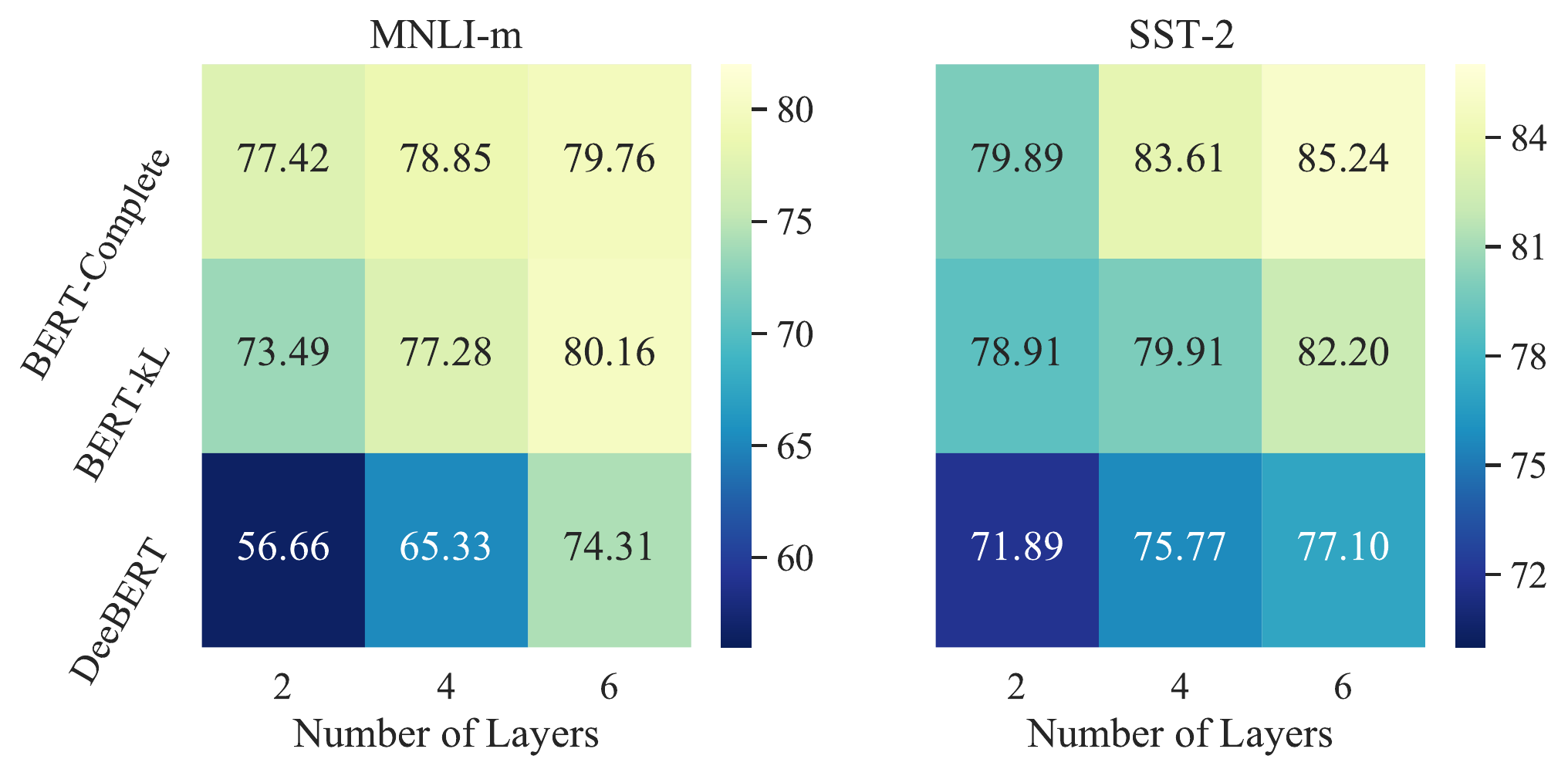}
    \caption{DIS~(\%, higher is better, see Eq.~\ref{eq:dis} in Section~\ref{subsec:dis}) heatmap of different models on the development set of MNLI and SST-2. 
    The DIS of internal off-ramps in the DeeBERT of shallow layers is lower than that of BERT-$k$l and BERT-Complete, which leads to more wrongly emitted instances.
    The exiting decisions in shallow layers of DeeBERT thus can be unreliable.}
    \label{fig:dis}
\end{figure}
We further probe whether the early exiting decisions made by internal classifiers are reliable, by first introducing two key concepts: 
\begin{itemize}
    \item \emph{Instance Difficulty} $d(x)$, which indicates whether an instance $x$ can be handled by a specific model. We define instances that the model cannot predict correctly as difficult instances, i.e., $d(x)=1$, and those can be handled well as easy ones, i.e., $d(x)=0$.
    \item \emph{Model Confidence} $c(x)$, which denotes how confident the model is about its prediction for a specific instance $x$. For each instance, we utilize the maximum class probability of the output distribution as the confidence score. 
\end{itemize}
% In more detail, we define instances that the model cannot predict correctly~(see Section~\ref{sec:difficulty} for details) as difficult instances, and those that can be handled well as easy ones.
Intuitively, a difficult instance should be predicted with less confidence than that of an easy one, such that the output distribution can be utilized as an indicator for early exiting decisions. However, the model confidence can be inconsistent with the instance difficulty due to the overconfident problem.
To measure this consistency, we propose \textit{Difficulty Inversion Score} (DIS).
Specifically, we first define a difficult inversion indicator function for instance pair $(x_i, x_j)$ measuring the inconsistency between model confidence and instance difficulty as:
\begin{equation}
\small 
    \text{DI}\left(x_{i}, x_{j}\right)=
    \left\{\begin{array}{ll}
1, & \text { if } d(x_i)> d(x_j) \text{ and } c(x_i) < c(x_j) \\
0, & \text { otherwise, }
\end{array}\right.
\end{equation}
% 进一步解释一下这个求和
The instances are then sorted by their confidence scores in an ascending order, i.e., $c(x_i) \leq c(x_j)$ for any $i <  j $. We compute the sum of difficulty inversion pair as:
\begin{equation}
    \text{DI-Sum} = \sum_{i=1}^{N} \sum_{j=1}^{i-1} \text{DI}( x_i, x_j),
\end{equation}
where $N$ is the number of instance.
The final $\text{DIS}$ is a normalized $\text{DI-Sum}$:
\begin{equation}
\label{eq:dis}
    \text{DIS} = 1 - \frac{1}{K} \ \text{DISum} ,
\end{equation}
where $K$ is a normalizing factor calculated as the product of the number of easy instances and the number of difficult instances, to re-scale DIS to the range from $0$ to $1$.
% A higher DIS indicates that the model performs well at ranking among instances according to the confidence score to distinguish difficult instances from easy ones.
According to the definition, the DIS measures the proportion of instance pairs that are correctly ranked by the classifier. Classifiers with lower DIS achieve lower consistency between the model confidence and instance difficulty, thus making more unreliable exiting decisions.
% , since harder instances are predicted with higher confidence, which results in emitting more wrongly predicted results.
The DIS thus can be utilized as a proxy for evaluating the quality of exiting decisions. 
% To measure the ability to rank the difficulty of instances, 
We compute the DIS on the development sets of MNLI-m and SST-2 for internal classifiers of different models discussed in Section~\ref{subsec:rep}, and the results are illustrated in Figure~\ref{fig:dis}. We find that:

(1) The DIS of internal classifiers in shallow layers of DeeBERT falls far behind BERT-$k$L and BERT-Complete. This indicates that the exiting decisions in the shallow layers of DeeBERT are unreliable, and the task performance thus can be poor when most instances are wrongly emitted in early layers.

(2) The ability to distinguish difficult examples from easy ones is enhanced as the layer number increases. 
Since the deep layer representations with semantic information can boost the task performance, it is reasonable to expect that the off-ramps in deep layers can provide more comprehensive early exiting decisions. 

Our analysis demonstrates that current dynamic early exiting predictions made by internal classifiers in shallow layers are not reliable, motivating us to inform the model of the instance difficulty for more robust exiting decisions. 
% Further analysis on the paragraph identification dataset QQP can be found in Appendix A, and the conclusions still hold.

% observations are consistent with that on the MNLI and SST-2 dataset.
% for more robust exiting decisions.
% We note that a simple temperature re-scaling technique have been applied to remedy the over-confident problem of internal off-ramps~\citep{schwartz-etal-2020-right}. However, since a normalizing factor for classification probability will not change the rank of instance probability, it cannot contribute to a better DIS metric.

% Researchers try to relieve this problem either by a post-hoc calibration via a temperature re-scaling factor learned based on the development set~\citep{schwartz-etal-2020-right} or directly enhancing the internal classifier via a more complicated predictor and self-distillation~\citep{Liu2020FastBERT}.

% Do not add the figure for now 
% \input{figures/cascade_model}

\section{Methodology}
\label{sec:method}
To remedy the drawbacks of conducting dynamic exiting in a single model, we extend the idea to a model cascade, by proposing CascadeBERT, that utilizes a suite of complete PLMs with different number of layers for acceleration in a cascading manner, and further devise a difficulty-aware calibration regularization to inform the model of instance difficulty.
\subsection{Cascade Exiting}
Formally, given $n$ complete pre-trained language models $\{M_1, \ldots, M_n\}$ fine-tuned on the downstream classification dataset, which are sorted in an ascending order by their corresponding number of layers $\{L_1, \ldots, L_n\}$, our goal is to conduct inference with the minimal computational cost for each input instance $x$ while maintaining the model performance. 
% 加上一句我们尝试过直接选取 model 的方案 但是比较困难
Our preliminary exploration shows that it is relatively hard to directly selecting a proper model for each instance according to the instance difficulty. 
Therefore, we formulate it as a cascade exiting problem, i.e., execute the model prediction sequentially for each input example from the smallest $M_1$ to the largest $M_n$, and 
check whether the prediction of the input instance $x$ can be emitted.
Specifically, we use the confidence score $c(x)$, i.e., the maximum class probability, as a metric to determine whether the predictions are confident enough for emitting:
\begin{equation}
    c(x) = \max_{y\in Y} (\Pr(y \mid x)),
\end{equation}
where $Y$ is the label set of the task and $\Pr(y \mid x)$ is the class probability distribution outputted by the current model.
The predicted result is emitted once the confidence score exceeds a preset threshold $\tau$. By varying the threshold $\tau$, we can obtain different speed-up ratios based on the application requirements. 
A smaller $\tau$ indicates that more examples are outputted using the current model, making the inference faster, while a bigger $\tau$ will make more examples go through larger models for better results.
% illustrated in Figure~\ref{fig:cascade} and
The cascaded exiting framework is summarized in Algorithm~\ref{alg:cascade}.
% 强调 Complete Cascade 的好处
Since every model in our cascading framework is a complete model, predictions are more accurate with instance representations that contain both low-level and high-level features, even when only the smallest model is executed.

\newcommand\mycommfont[1]{\footnotesize\ttfamily{#1}}
\SetCommentSty{mycommfont}
\SetAlCapFnt{\small}
\SetAlCapNameFnt{\small}
\SetKwInput{KwInput}{Input}
\begin{algorithm}[t]
\small 
\DontPrintSemicolon
\KwInput{Models $\{ M_1, \ldots, M_n\}$, threshold $\tau$}
\KwData{Input $x$}
\KwResult{Class probability distribution $\Pr(y \mid x)$}
\For{$i \gets 1$ \KwTo $n$ }{
  \tcp{calculate class distribution}
  $\Pr(y|x) = M_i(x)$ \;
  \tcp{compute confidence score}
%   \Comment calculate class distribution \;
  $c(x) = \max_y(\Pr(y\mid x))$ \;
%   \Comment compute confidence score\;
  \If{$c(x) > \tau$}{
   Early exit $\Pr(y \mid x)$\;
    }
}
\Return $\Pr(y \mid x)$
\caption[Cascade]{Cascade Exiting}
\label{alg:cascade}
\end{algorithm}

\subsection{Difficulty-Aware Regularization}
\label{sec:difficulty}
% We can further reduce the overhead cost brought by re-running the small models by directly selecting proper model to go through.
To further make the cascade exiting based on confidence score more reliable, we design a difficulty-aware regularization~(DAR) objective based on instance difficulty, to regularize the model classifiers produce lower confidence for more difficult instances.
To measure the instance difficulty, we first split the training dataset $\mathcal{D}$ into $K$ folds $\{\Tilde{\mathcal{D}}_i \mid i = 1, \ldots, K \}$.
For each complete model in our cascade, we train $K$ models with the same architecture using the leave-one-out method, e.g., model $M_n^i$ is trained on the $\mathcal{D}_{- \Tilde{\mathcal{D}}_i} = \bigcup_j^{j \neq i}  \Tilde{\mathcal{D}}_j$. 
We utilize $M_n^i$ to evaluate the difficulty of the examples in $\Tilde{\mathcal{D}}_i$ for model $M_n$. 
Specifically, the samples are marked as easy, i.e., $d=0$, if they can be correctly classified by the model. Otherwise, they are marked as difficult, i.e., $d=1$.
% The difficulty of examples in $\Tilde{D}_i$
% is evaluated by $\theta_i$, by labeling the correctly predicted examples as easy and wrongly predicted ones as difficult. 
% The training examples are predicted by the models that is not trained on it.
To eliminate the impact of randomness, we group the predictions of $5$ seeds and strictly label the examples that can be correctly predicted in all seeds as easy examples, while the others as difficult ones.

% of the cascade mechanism.
With the instance difficulty for each instance in the training dataset, we add a difficulty-based margin objective for each instance pair:
\begin{equation}
\small{
    \mathcal{L}(x_i, x_j) = \max \left\{ 0,   -g\left(x_i, x_j\right) \left(c\left(x_i\right) - c\left(x_j\right)\right)  + \epsilon\right\},
}
\end{equation}
where $\epsilon$ is a confidence margin. We design $g\left(x_i, x_j\right)$ as below:
\begin{equation}
    g\left(x_{i}, x_{j}\right)=\left\{\begin{array}{ll}
1, & \text { if } d(x_i)>d(x_j) \\
0, & \text { if } d(x_i)=d(x_j) \\
-1, & \text {otherwise}.
\end{array}\right.
\end{equation}
This objective is added to the original task-specific loss with a weight factor $\lambda$ to adjust its impact.
% \footnote{We construct pairs in one mini-batch, and the result is a sum of all pairs.}
% 写一下怎么计算 这个 L(x_i, x_j) 在mini-batch 之中
By optimizing the above objective function, the confidence scores of difficult instances are adjusted to be lower than those of easy instances, thus making the confidence-based emitting decisions more reliable.
% Note that traditional post-hoc calibrated methods like temperature scaling~\citep{guo2017calibration} are not applicable in this case, since the re-scaling technique will not change the rank for different instances.

% To measure the instance difficulty, we first split the training dataset $\mathcal{D}$ into $K$ folds $\{\Tilde{\mathcal{D}}_i \mid i = 1, \ldots, K \}$.
% For each complete model in our cascade, we train $K$ models using the leave-one-out method, i.e., model $M_n^i$ is trained on the $\mathcal{D}_{- \Tilde{\mathcal{D}}_i} = \bigcup_j^{j\neq i}  \Tilde{\mathcal{D}}_j$. 
% We utilize $M_n^i$ to evaluate the difficulty of the examples in $\Tilde{\mathcal{D}}_i$. Specifically, the samples are marked as easy examples~($d=0$) if they can be correctly classified by the model. Otherwise, they are marked as difficult~($d=1$).
% % The difficulty of examples in $\Tilde{D}_i$
% % is evaluated by $\theta_i$, by labeling the correctly predicted examples as easy and wrongly predicted ones as difficult. 
% % The training examples are predicted by the models that is not trained on it.
% To eliminate the impact of randomness, we group the predictions of $5$ seeds and strictly label the examples that can be correctly predicted in all seeds as easy examples, while the others as difficult ones.
% A similar approach is adopted by \citet{xu2020curriculum} for applying curriculum learning into natural language understanding.

% For example, the input first go through a $2$-layer model and
\section{Experiments}
We evaluate our method on the classification tasks in the GLUE benchmark~\citep{wang-etal-2018-glue} with BERT~\citep{devlin2019bert}. 
We first give a brief introduction of the dataset used and the experimental setting, followed by the description of baseline models for comprehensive evaluation. The results and analysis of the experiments are presented last.
\subsection{Experimental Settings}

We use six classification tasks in GLUE benchmark, including MNLI~\citep{williams2018mnli}, MRPC~\citep{dolan2005mrpc}, QNLI~\citep{rajpurkar2016squad}, QQP,\footnote{\url{https://data.quora.com/First-Quora-Dataset-Release-Question-Pairs}} RTE~\citep{bentivogli2009rte} and SST-2~\citep{socher2013sst}. 
The metrics for evaluation are F1-score for QQP and MRPC, and accuracy for the rest tasks. 
Our implementation is based on the Huggingface Transformers library~\citep{wolf-etal-2020-transformers}.
We use two models for selection with $2$ and $12$ layers, respectively, since they can provide a wide range for acceleration. 
The difficulty score is thus evaluated based on the $2$-layer model.
The effect of incorporating more models in our cascade framework is explored in the later section.
% is sufficient for achieving a smooth enough speed-up and performance and trade-off curve.
We utilize the weights provided by~\citet{turc2019well} to initialize the models in our suite. The split number $K$ for difficulty labeling is set to $8$.
% since they are trained from scratch to obtain the full pipeline processing ability.
We use AdamW~\citep{loshchilov2018adamw} with a learning rate $2\text{e-}5$ to train model for $10$ epochs, since we find that small models need more time to converge.
We set DAR weight $\lambda$ as $0.5$, and perform grid search over $\epsilon$ in $\{0.1, 0.3, 0.5, 0.7\}$. The best model is selected based on the validation performance.
The statistics of datasets and the selected $\epsilon$ are provided in Table~\ref{tab:dataset}.

\begin{table}[t!]
    \centering
    \small
    \begin{tabular}{@{}lrrrcc@{}}
    \toprule
        \textbf{Dataset} & \textbf{\# Train} &\textbf{\# Dev} & \textbf{\# Test} & \textbf{Metric}  &  $\epsilon$\\
        \midrule
         MNLI &  393k    &20k &20k  &  Accuracy & 0.3  \\ 
         MRPC &   3.7k  & 0.4k & 1.7k &  F1-score & 0.5 \\ 
        QNLI  &   105k  & 5.5k & 5.5k &  Accuracy &0.3  \\ 
          QQP &  364k   & 40k& 391k& F1-score & 0.3 \\ 
          RTE &   2.5k  & 0.3k &3k  &  Accuracy &0.5\\ 
         SST-2 & 67k   & 0.9k & 1.8k&  Accuracy  & 0.5 \\ 
         \bottomrule
    \end{tabular}
    \caption{Statistics of six classification datasets in GLUE benchmark. The selected difficulty margins $\epsilon$ of each datasets are provided in the last column. }
    \label{tab:dataset}
\end{table}

\begin{table*}[t]
    \centering
    \small
    \setlength{\tabcolsep}{4pt}
    \begin{tabular}{@{}l@{\hspace{2pt}}l|cccccc|c@{}}
    \toprule 
        \multicolumn{2}{c|}{\textbf{Method}} & \textbf{MNLI-m}/\textbf{mm} & \textbf{MRPC} & \textbf{QNLI}& \textbf{QQP}& \textbf{RTE} &\textbf{SST-2} & \textbf{AVG} \\
    \midrule 
     & BERT-base\textsuperscript{\textdagger}   & 84.6 {\scriptsize(1.00$\times$)} / 83.4 {\scriptsize(1.00$\times$)}  & 88.9 {\scriptsize(1.00$\times$)} & 90.5 {\scriptsize(1.00$\times$)} & 71.2 {\scriptsize(1.00$\times$)} & 66.4 {\scriptsize(1.00$\times$)} &  93.5 {\scriptsize(1.00$\times$)} &  82.6  \\
    \midrule
        \multirow{4}{*}[-3pt]{\rotatebox[origin=c]{90}{$\sim$\ 2$\times$}}  & BERT-6L\textsuperscript{\textdaggerdbl}  & 79.9 {\scriptsize(2.00$\times$)} / 79.2 {\scriptsize(2.00$\times$)} & 85.1 {\scriptsize(2.00$\times$)} & 86.2 {\scriptsize(2.00$\times$)}  &68.9 {\scriptsize(2.00$\times$)} & \textbf{65.0} {\scriptsize(2.00$\times$)} & 90.9 {\scriptsize(2.00$\times$)} &  79.3\\
     & DeeBERT\textsuperscript{\textdagger}   & 74.4 {\scriptsize(1.87$\times$)} /  73.1 {\scriptsize(1.88$\times$)} & 84.4 {\scriptsize(2.07$\times$)}  & 85.6 {\scriptsize(2.09$\times$)} & 70.4 {\scriptsize(2.13$\times$)}  &  64.3 {\scriptsize(1.95$\times$)} & 90.2 {\scriptsize(2.00$\times$)}  &  77.5 \\
     & PABEE\textsuperscript{\textdagger}    &  79.8 {\scriptsize(2.07$\times$)} / 78.7 {\scriptsize(2.08$\times$)} &84.4 {\scriptsize(2.01$\times$)}  & 88.0 {\scriptsize(1.87$\times$)}& 70.4 {\scriptsize(2.09$\times$)} &  64.0 {\scriptsize(1.81$\times$)}& 89.3 {\scriptsize(1.95$\times$)} & 79.2\\
         & \textbf{CascadeBERT}  & \textbf{83.0} {\scriptsize(2.01$\times$)} / \textbf{81.6} {\scriptsize(2.01$\times$)}& \textbf{85.9} {\scriptsize(2.01$\times$)} &  \textbf{89.4} {\scriptsize(2.01$\times$)} & \textbf{71.2} {\scriptsize(2.01$\times$)}  & 64.6 {\scriptsize(2.03$\times$)} &  \textbf{91.7} {\scriptsize(2.08$\times$)}   & \textbf{81.1}  \\
    \midrule 
    \multirow{4}{*}[-3pt]{\rotatebox[origin=c]{90}{$\sim$\ 3$\times$}}  & BERT-4L\textsuperscript{\textdaggerdbl}  & 75.8 {\scriptsize(3.00$\times$)} / 75.1 {\scriptsize(3.00$\times$)} & 82.7 {\scriptsize(3.00$\times$)} & 84.7 {\scriptsize(3.00$\times$)}  &66.5 {\scriptsize(3.00$\times$)} & 63.0 {\scriptsize(3.00$\times$)} & 87.5 {\scriptsize(3.00$\times$)} & 76.5 \\
     & DeeBERT\textsuperscript{\textdaggerdbl}   & 63.2 {\scriptsize(2.98$\times$)}  / 61.3 {\scriptsize(3.03$\times$)}  & 83.5 {\scriptsize(3.00$\times$)}  & 82.4 {\scriptsize(2.99$\times$)} & 67.0 {\scriptsize(2.97$\times$)}  &  59.9 {\scriptsize(3.00$\times$)} & 88.8 {\scriptsize(2.97$\times$)}   &  72.3\\
& PABEE\textsuperscript{\textdaggerdbl}  & 75.9 {\scriptsize(2.70$\times$)} / 75.3 {\scriptsize(2.71$\times$)}& 82.6 {\scriptsize(2.72$\times$)} &  82.6 {\scriptsize(3.04$\times$)}& 69.5 {\scriptsize(2.57$\times$)} &  60.5 {\scriptsize(2.38$\times$)} &85.2 {\scriptsize(3.15$\times$)} & 75.9\\ 
 & \textbf{CascadeBERT}  & \textbf{81.2} {\scriptsize(3.00$\times$)} /  \textbf{79.5} {\scriptsize(3.00$\times$)} &  \textbf{84.0} {\scriptsize(3.00$\times$)} &   \textbf{88.5} {\scriptsize(2.99$\times$)} &  \textbf{71.0} {\scriptsize(3.02$\times$)}  &   \textbf{63.8} {\scriptsize(3.03$\times$)} &  \textbf{90.9} {\scriptsize(2.99$\times$)}   &  \textbf{79.8} \\
 \midrule 
     \multirow{4}{*}[-3pt]{\rotatebox[origin=c]{90}{$\sim$\ 4$\times$}}  & BERT-3L\textsuperscript{\textdaggerdbl}  & 74.8 {\scriptsize(4.00$\times$)} / 74.3 {\scriptsize(4.00$\times$)} & 80.5 {\scriptsize(4.00$\times$)} & 83.1 {\scriptsize(4.00$\times$)}  &65.8 {\scriptsize(4.00$\times$)} &  55.2 {\scriptsize(4.00$\times$)} &  86.4 {\scriptsize(4.00$\times$)} &  74.3 \\ % From Liao Kaiyuan 
     & DeeBERT\textsuperscript{\textdaggerdbl}   &  55.8 {\scriptsize(4.01$\times$)}  / 54.2 {\scriptsize(3.99$\times$)}  &  \textbf{82.9} {\scriptsize(3.99$\times$)}  & 75.9 {\scriptsize(4.00$\times$)} &  62.9 {\scriptsize(4.01$\times$)}  &  57.4 {\scriptsize(4.00$\times$)} &  85.4 {\scriptsize(4.00$\times$)}   & 67.8 \\
& PABEE\textsuperscript{\textdaggerdbl}  & 62.3 {\scriptsize(4.32$\times$)} / 63.0  {\scriptsize(4.30$\times$)}& 79.9 {\scriptsize(4.00$\times$)} &  - & 68.0 {\scriptsize(3.45$\times$)} & 56.0 {\scriptsize(3.62$\times$)} & - & -\\  % From Liao Kaiyuan
 & \textbf{CascadeBERT}  & \textbf{79.3} {\scriptsize(4.03$\times$)} /  \textbf{77.9} {\scriptsize(3.99$\times$)} &  82.6 {\scriptsize(4.00$\times$)} &   \textbf{86.5} {\scriptsize(3.99$\times$)} &  \textbf{70.0} {\scriptsize(4.04$\times$)}  &   \textbf{61.6} {\scriptsize(4.02$\times$)} &  \textbf{90.3} {\scriptsize(4.01$\times$)}   &  \textbf{78.3} \\
    \bottomrule
    \end{tabular}
    \caption{Test results from the GLUE server. We report F1-score for QQP and MRPC and accuracy for other tasks, with the corresponding speed-up ratios shown in parentheses. For baseline methods,
    \textsuperscript{\textdagger} denotes results taken from the original paper and \textsuperscript{\textdaggerdbl} denotes results based on our implementation.
    The - denotes that results are not available by tuning the threshold of PABEE.
    % The upper, middle and bottom rows report model performance around 2$\times$, 3$\times$ and 4$\times$ acceleration, respectively. 
    Best results are shown in \textbf{bold}.}
    \label{tab:glue_test}
\end{table*}

The inference speed-up ratio is estimated as the ratio of number of the original model and layers actually executed in forward propagation in our cascade. 
% , as the inference time comparison can be unstable due to different hardware and implementations
Compared to performing dynamic exiting in a single model, the overhead of CascadeBERT consists of two parts. 
The former is the extra embedding operations, which is nearly $0.3$M FLOPs and is negligible compared with the $1809.9$M FLOPs of each layer~\citep{Liu2020FastBERT}.
The latter is brought by instances that run forward propagation multiple times, which is counted in the speed-up ratio calculation. 
For example, for an instance which is first fed into a $2$-layer model and then goes through a $4$-layer model to obtain the final prediction result, the number of layers actually executed is therefore $6$ and the corresponding speed-up ratio is $2\times$ compared to the original $12$-layer full model.
% The actual inference time comparison in Appendix C shows the speed-up ratio estimation is reasonable.
% 这个数可以是小于 1 的
% Due to the overhead, the speed-up ratio of our method can be less than $1$. We can tackle this in practice by executing the cascade models in parallel as models are independent with each other.
% The execution is stopped once a credible result is obtained, and the speed-up ratio can thus have a rough lower bound $1\times$.

% Similar metrics are also adopted by~\citet{Xin2020DeeBERT} and \citet{}.
% of layer that examples go through during inference, due to the measurement of run-time differs from different hardwares and implementations. We manually adjust the exit threshold $\tau$ and calculate the speed-up ratio by comparing the actually executed layers in forward propagation and the original required layers 
% Specifically, the speed-up ratio over an original model with $n$-layer is calculated as:
% \begin{equation}
%     \text{speed-up ratio} = \frac{\sum_{i=1}^N n \times m^i}{\sum_{i=1}^N  C \times m^i}
% \end{equation}
% where $m^i$ is the instance number that actually costs $C$ layers in total and $N$ is the number of test instances.

\subsection{Baselines}
We implement two kinds of baselines, including:

\medskip
\noindent\textbf{Early Exiting}, including BERT-$k$L, where the first $k$ layers with a fine-tuned classifier are used for outputting the final classification results. We take $k=6$, $k=4$ and $k=3$ to obtain a statically compressed model with speed-up ratios of 2$\times$, 3$\times$ and $4\times$, respectively;
DeeBERT~\citep{Xin2020DeeBERT}, which makes dynamic early predictions based on the internal classifiers;
PABEE~\citep{zhou2020pabee}, an enhanced variant by emitting the result until several layers produce a consistent prediction.~\footnote{PABEE provides limited speed-ratios since the threshold for tuning speed-up ratios can only be set to integers.}
% Note that.

\medskip
\noindent\textbf{Knowledge Distillation} methods that do not require external data, including DistilBERT~\citep{Sanh2019DistilBERT}, which distills knowledge from the teacher model to the student during pre-training via logit distillation; BERT-PKD~\citep{Sun2019PatientKD}, which distills internal states of the teacher model to the student model; BERT-of-Theseus~\cite{Xu2020BERTofTheseus}, which gradually replaces the module in the original model;  BERT-PD~\citep{turc2019well}, which directly pre-trains a compact model from scratch and conducts distillation on the task dataset.
% Note that these distillation techniques can be further incorporated into our framework to enhance the performance.

\begin{table}[t]
\centering
\small
\setlength{\tabcolsep}{2.5pt}
\begin{tabular}{@{}l|cccc|c@{}}
\toprule
{\textbf{Method}} & \textbf{MNLI-m}/\textbf{mm}  &  \textbf{QNLI} & \textbf{QQP} & \textbf{SST-2} & \textbf{AVG} \\
\midrule 
DistilBERT &  78.9 / 78.0  &  85.2  &  68.5 & \textbf{91.4}  & 80.4\\  % From TinyBERT
BERT-PKD & 79.9 / 79.3  &  85.1 & 70.2  & 89.4 &  80.8\\  % From TinyBERT
BERT-Theseus & 78.6 /  77.4  &  85.5  & 68.3 &   89.7 & 79.9\\ 
BERT-PD & 79.3 / 78.3 & 87.0 & 69.8 & 89.8 &  80.8\\ 
% TinyBERT &82.5 / 81.8 & 87.7 & 71.3 & 92.6 & 83.2  \\ 
CascadeBERT & \textbf{81.2} / \textbf{79.5}   & \textbf{88.5}   & \textbf{71.0}   & 90.9  & \textbf{82.2} \\ 
\bottomrule
\end{tabular}
\caption{
Test result comparison with static knowledge distillation methods under speed-up ratio $3\times$. 
% \textsuperscript{\textdagger} denotes results taken from \citet{Jiao2019TinyBERT} and \textsuperscript{\textdaggerdbl} are results from the corresponding original paper.
}
\label{tab:kd_result}
\end{table}

\subsection{Overall Results}
The performance comparison with early exiting methods are presented in Table~\ref{tab:glue_test}.
% We surprisingly find that the early exiting method cannot beat the simple method BERT-\textit{n}L, which directly fine-tunes a classifier layer after an internal layer. This validates our motivation that the emitting decisions and the predictions based on shallow layer representations are not reliable.
% Our method instead makes a selection between complete models in a cascading manner and achieves superior performance over early exiting methods with a large margin. 
% Furthermore, our CascadeBERT also outperforms the enhanced version of dynamic early exiting, PABEE, by a $0.9$ average points when the speed-up ratio is 2$\times$.
% The gap becomes clearer when the speed-up ratio comes to 3$\times$, which demonstrates that our model can maintain satisfying performance even when the speed-up ratio is relatively high. 
 % 现象 + 结论
 % 现象1： 超过所有 baseline -> 验证方法整体的有效性
We observe that CacadeBERT outperforms all the baseline methods under different speed-up ratios, validating the effectiveness of our proposal.
% 现象2： 高加速比差距更明显 -> complete representations 的有效性
Furthermore, the performance gap becomes clearer as the acceleration ratio increases. For example, CascadeBERT outperforms DeeBERT by a big margin with a relative 15.5\% improvement~(10.5 points on average) under speed-up ratio $4\times$.
This phenomenon demonstrates that CascadeBERT can break the performance bottleneck by utilizing comprehensive representations from complete models.
Interestingly, we find CascadeBERT performs closely with DeeBERT on MRPC.
We attribute it to that this paraphrase identification task requires less high-level semantic information, thus only utilizing low-level features at specific layers can sometimes become beneficial. 
% FastBERT 单独说
Different from DeeBERT and PABEE, FastBERT~\citep{Liu2020FastBERT} enhances the internal classifiers with a self-attention mechanism to use all the hidden states for predictions, resulting in a different magnitude of computational overhead. 
Comparison results with FastBERT are provided in Appendix~\ref{apx:fastbert}. CascadeBERT can still outperform FastBERT, especially on the tasks requiring semantic reasoning ability. 

% e.g., a $3.8$ average points over the strong PABEE baseline under speed-up ratio $3\times$, demonstrating the superiority of CascadeBERT. 

% 如果这里的结论是 12L 的 pipeline 无法用 KD 压缩到 4L -> 这样就打自己的脸了...退出 BERT-PD 也没有 pipeline 
% 应该是不同 实例需要不同粒度的 pipeline 能力 我们的可以提供 flexible 的选择
% 现象： 超过 KD 方法
Besides, our proposal also achieves superior performance over strong knowledge distillation methods like BERT-PKD and BERT-of-Theseus, as shown in Table~\ref{tab:kd_result}.
% Although distillation methods can implicitly learn the pipeline ability by forcing student models to mimic the behaviors of the teacher model, 
% % it is still relatively hard to obtain models with pipeline ability of different granularity for different instances.
% it is still relatively hard to obtain a highly compressed student model with the pipeline processing ability. 
% 结论： 单独的一个 model 无法处理不同 instance，
Distillation methods can implicitly learn the semantic reasoning ability by forcing student models to mimic the behaviors of the teacher model. However, it is still relatively hard to obtain a single compressed model to handle all instances well, as different instances may require the reasoning ability of different granularities.
Our cascading mechanism instead provides flexible options for instances with different complexities, thus achieving better results.
% since every model candidate in our framework is a complete model and the predictions are calibrated, the cascade of different models can still produce robust results.

% \begin{figure}[t]
% \centering
% \subfigure[DeeBERT]{
% \begin{minipage}[b]{0.4\textwidth}
% \includegraphics[width=\linewidth]{imgs/mnli-deebert-2L-tsne.pdf}
% \end{minipage}
% }
% \subfigure[BERT-Complete]{
% \begin{minipage}[b]{0.4\textwidth}
% \includegraphics[width=\linewidth]{imgs/mnli-complete-2L-tsne.pdf}
% \end{minipage}
% }
% \caption{t-SNE visualization} \label{fig:tsne}
% \end{figure}

\begin{figure}[t]
\centering
\subfigure[Instance representations t-SNE projection on MNLI-m.]{
\begin{minipage}[t]{0.95\linewidth}
\centering
\includegraphics[width=\linewidth]{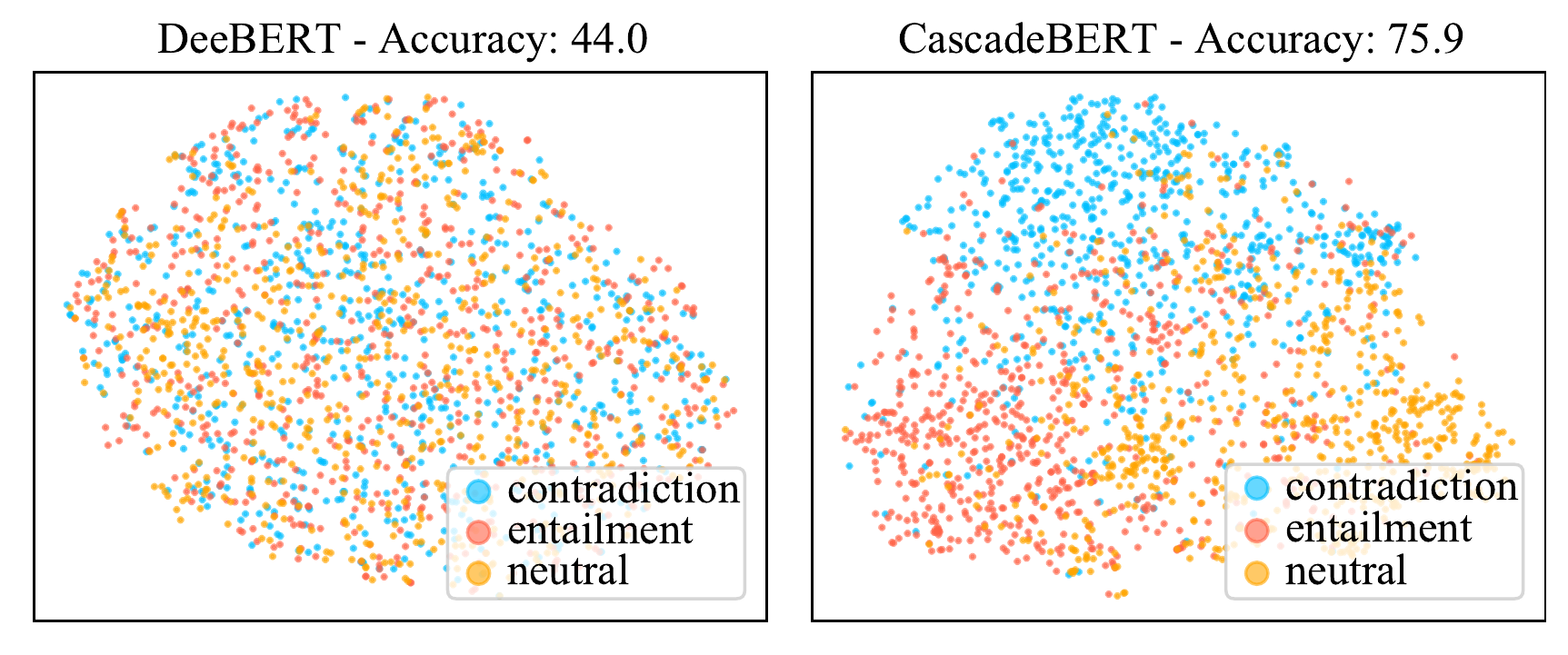}
 \label{fig:tsne_deebert}
\end{minipage}%
}%
\quad
\subfigure[Instance representations t-SNE projection on SST-2.]{
\begin{minipage}[t]{0.95\linewidth}
\centering
\includegraphics[width=\linewidth]{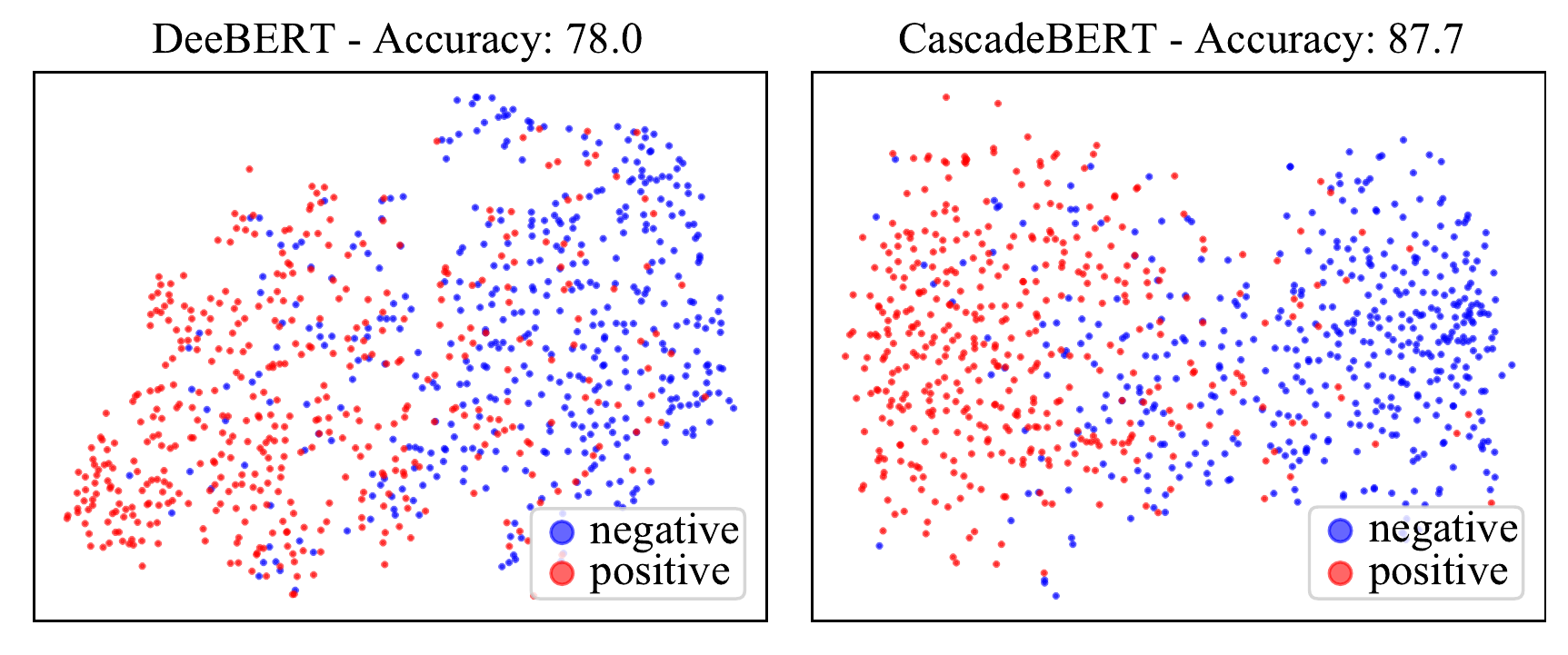}
%\subcaption{fig1}
\end{minipage}%
\label{fig:tsne_complete}
}%
\centering
\caption{t-SNE visualization of instance representations of different class in DeeBERT and our CascadeBERT at the second layer. 
The instance representations of our CascadeBERT exhibit a more distinct boundary between different classes, helping the following classifier to make accurate predictions. Best viewed in color.}
\label{fig:tsne}
\end{figure}
\section{Analysis}
In this section, we investigate how the proposed CascadeBERT makes accurate predictions under high speed-up ratios, and analyze the effects of the proposed difficulty-aware regularization and incorporating more models to the cascade. We finally examine the generalizability by applying it to RoBERTa.
The experiments are conducted on MNLI, QNLI, QQP and SST-2 for stable results.
% The experiments are conducted on MNLI, QNLI, QQP and SST-2 since their data sizes are relatively large for obtaining stable results.
\subsection{Visualization of Instance Representations}
To investigate how the representations with sufficient information benefit accurate predictions, we visualize the instance representations after 2 layers
% that will be fed into the classifier 
using t-SNE projection~\citep{maaten2008tsne}.
The results and the corresponding classifier accuracy are shown in Figure~\ref{fig:tsne}.
We observe that the boundary of instances belonging to different classes of our CascadeBERT is much clearer than that of DeeBERT.
Since the representations contain sufficient information for predictions, our model can thus obtain more accurate results.
% 说一下 MNLi 和 SST2的差别 反应了任务的难度
Interestingly, the shallow representations in DeeBERT of SST-2 are already separable to some extent, which indicates that the task is somewhat easy. It is consistent with our main results that the performance degradation of different methods is negligible on SST-2. 
% Calibration 的效果
\subsection{Effects of Difficulty-Aware Regularization}
% \noindent\textbf{Difficulty-aware regularization improves the performance and speed-up trade-off curve}.
We show the performance of an ablated version of our proposal, CascadeBERT w/o DAR in Table~\ref{tab:ablation}. 
% The results indicate that the DAR can make the predictions better reflect whether current model can handle the example, thus improving the performance of our framework. 
The results indicate that the DAR can improve the overall performance of our framework. 
Note that the improvement is very challenging to achieve, as the original model cascade already outperforms strong baseline models like PABEE.
% 输出 difficulty 的 confidence score & DIS 
% Furthermore, we explore whether the proposed DAR can help the model to distinguish difficult instances from easy ones better.
Furthermore, we explore whether the performance boost comes from an enhanced ability of the model to distinguish difficult instances from easy ones.
Specifically, we compute the DIS and the task accuracy~(Acc) of the smallest model in our cascade. The results are listed in Table~\ref{tab:dis_score}.
\begin{table}[t]
\centering
\small
\setlength{\tabcolsep}{2pt}
\begin{tabular}{@{}l@{\hspace{2pt}}l|cccc|c@{}}
\toprule
\multicolumn{2}{c|}{\textbf{Method}} & \textbf{MNLI-m}/\textbf{mm}  &  \textbf{QNLI} & \textbf{QQP} & \textbf{SST-2} & \textbf{AVG} \\
% \midrule 
% \multirow{2}{*}{\rotatebox[origin=c]{90}{$\sim$2$\times$}}  & CascadeBERT & 83.0 / 81.6 & 89.4  &  71.2 & 91.7 &  83.4 \\ 
% & ~~ -  w/o DAR & 82.0 / 81.7 & 88.8 &  71.0& 91.9&  83.1\\ 
\midrule 
\multirow{2}{*}{\rotatebox[origin=c]{90}{$\sim$3$\times$}}&CascadeBERT &  \textbf{81.2} / \textbf{79.5}  & \textbf{88.5} & \textbf{71.0} & \textbf{90.9} & \textbf{82.2} \\ 
& ~~ - w/o DAR & 80.0 / 79.3 & 87.8 &\textbf{71.0} &  90.3&   81.7\\ 
\midrule 
\multirow{2}{*}{\rotatebox[origin=c]{90}{$\sim$4$\times$}}&CascadeBERT &  \textbf{79.3} / 77.9  & 86.5 & \textbf{70.0} & \textbf{90.3} & \textbf{80.8} \\ 
& ~~ - w/o DAR & 78.9 / \textbf{78.1} & \textbf{86.6} &69.8 &  89.6&   80.6\\ 
\bottomrule
\end{tabular}
\caption{Ablated results of the proposed difficulty-aware regularization under different speed-up ratios.
}\label{tab:ablation}

\end{table}
% 现象： DAR 提升 DIS，说明 给模型 hint 更可靠的退出决策
We find that the DAR can effectively improve the DIS while slightly harms the task performance, indicating that DAR boosts the overall performance by helping model make more reliable emitting decisions.
The exceptional decrease of DIS on QQP is attributed to the fact that the original DIS score is relatively high, which makes further improvements very challenging. 
Besides, the DAR can lower the prediction confidence of difficult instances, which improves the consistency between the predicted probability and how likely the model is to be correct for an instance.
We quantitively measure this calibration effect of DAR, by utilizing the expected calibration error~(ECE)~~\citep{guo2017calibration}.\footnote{Refer to Appendix~\ref{apx:ece} for the details of the ECE score.}
As shown in Table~\ref{tab:dis_score}, the DAR not only improves the DIS score, but also calibrates the model predictions, achieving lower expected calibration error.  % 同时 ECE 下降

\begin{table}[t]
\centering
\small 
\begin{tabular}{@{}lc|rrr@{}}
\toprule
\textbf{Dataset}     & \textbf{Method} &  \textbf{DIS} ($\uparrow$)& \textbf{Acc}($\uparrow$) &  \textbf{ECE} ($\downarrow$) \\
\midrule
\multirow{2}{*}{MNLI-m} & CascadeBERT &   \textbf{78.00}  & 75.97  &   \textbf{7.90} \\
    & - w/o DAR    &  76.73  & \textbf{76.02} & 11.07   \\
\midrule
\multirow{2}{*}{QNLI} &CascadeBERT   &   \textbf{78.89}  & 84.53 & \textbf{3.41}\\
       & - w/o DAR   &   77.79   & \textbf{84.73} &  8.79  \\
\midrule
\multirow{2}{*}{QQP} & CascadeBERT  &   84.39     & 87.21 & \textbf{3.37}\\
    & - w/o DAR   &  \textbf{85.77}  &\textbf{88.71} & 4.99 \\
\midrule 
\multirow{2}{*}{SST-2} & CascadeBERT &  \textbf{82.02}& 87.70 & \textbf{5.61} \\
    & - w/o DAR   &  79.30   &\textbf{87.95}  &    8.73 \\
\bottomrule
\end{tabular}
\caption{The ECE (\%), Acc (\%) and DIS (\%) scores on different datasets.
$\uparrow$ denotes higher is better, while $\downarrow$ means lower is better. 
The proposed DAR can boost the performance by giving hints of instance difficulty and calibrate the model predictions. }
\label{tab:dis_score}
\end{table}

% QNLI
% w/o dar: 0.8473366282262493
% w/ dar:  0.8453230825553725

% QQP
% w/o dar: 0.887088795448924
% w/ dar: 0.8720504575810042

% SST2
% w/o dar: 0.8795871559633027
% w/ dar:  0.8623853211009175 / 0.8769724770642202

% MNLI
% w/o dar: 0.7602649006622516
% w/  dar: 0.7596535914416709
% 多模型的结果 以及 理论的收益分析
\subsection{Impacts of More Models in Cascade}
\begin{figure}[t]
    \centering
    \includegraphics[width=0.95\linewidth]{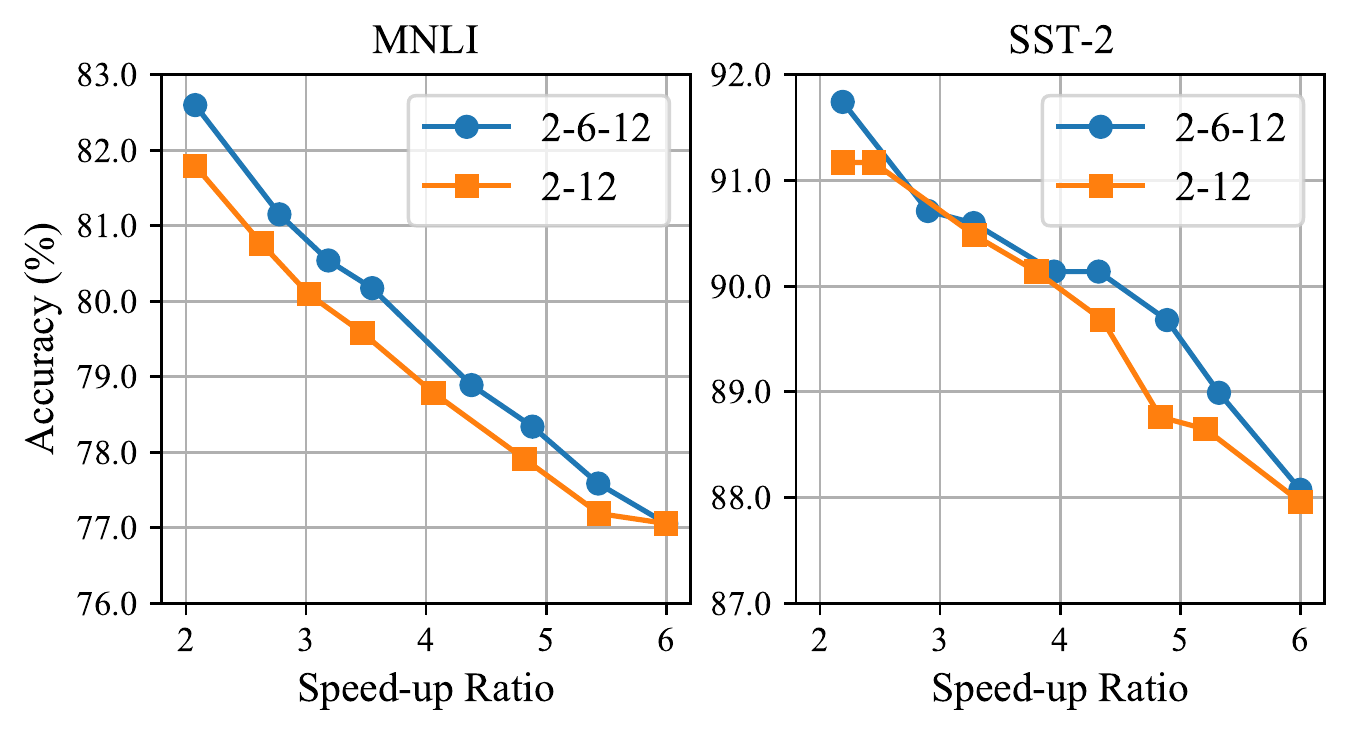}
    \caption{Task performance on the validation set and speed-up ratio trade-off curve comparison of a 2-model cascade~(orange square) and a 3-model cascade~(blue circle) on SST-2 and MNLI-m.}
    \label{fig:more_cascade}
\end{figure}

We further consider to incorporate more models into the CascadeBERT framework. 
Theoretically, we prove that adding more models in cascade can boost the task performance under mild assumptions. Besides, the benefits will become marginal as the number of model increases.
% , under mild assumptions about the performance of the incorporated model and instances exiting distribution. 
The detailed proof is provided in the Appendix~\ref{apx:theory_more_teacher}.
We empirically verify this by adding a medium-sized model with $6$ layers which satisfies our assumptions into the cascade. 
The performance under different speed-up ratios of a 2-12 cascade consists of a $2$L model and a $12$L model and the above mentioned 2-6-12 cascade are illustrated in Figure~\ref{fig:more_cascade}.
% We vary the exiting thresholds to evaluate the task performance under different speed-up of a 2-12 cascade consists of a $2$L model and a $12$L model and the above mentioned 2-6-12 cascade.
% The results are illustrated in Figure~\ref{fig:more_cascade}.
Overall, we find that adding a model with a moderate size can slightly improve the performance, while the gain becomes marginal when the speed-up ratio is higher, since most instances are emitted from the smallest model.

% 加一节 ALBERT-2L & 12L / RoBERTa 2L -12L 证明 
% 1. 没有 Pipeline 模型的替代 task-specific model
% 2. 框架的通用性

\subsection{Fine-tuned Models as an Alternative}
To verify the generalizability of our cascading framework, we propose to apply our method to RoBERTa~\citep{Liu2019RoBERTa}.
However, small versions of RoBERTa pre-trained from scratch are currently not available. 
We notice that BERT-$k$L model can achieve comparable performance via fine-tuning, as discussed in Section~\ref{sec:analysis}. 
Therefore, we propose to leverage a fine-tuned RoBERTa-2L with the vanilla KD~\citep{Hinton2015Distilling} incorporated for enhancing its semantic reasoning ability, as an alternative of the original complete model.
% Specifically, a logit matching loss is added when fine-tuning the cascade models.
The results around $3\times$ speed-up are listed in Table~\ref{tab:roberta}.
Our framework still outperforms dynamic early exiting baselines by a clear margin, validating that our framework is universal and can be combined with knowledge distillation techniques to further boost the performance.

\begin{table}[t]
\centering
\small
\setlength{\tabcolsep}{1.5pt}
\begin{tabular}{@{}l|cccc|c@{}}
\toprule
{\textbf{Method}} & \textbf{MNLI-m}/\textbf{mm}  &  \textbf{QNLI} & \textbf{QQP} & \textbf{SST-2} & \textbf{AVG} \\
\midrule 
RoBERTa-base & 87.0 / 86.3 &92.4  &  71.8 &94.3 &  86.4\\ 
\midrule
RoBERTa-4L &  \textbf{80.3} / \textbf{79.2}  & 86.2 & 69.8 & 90.8 & 81.2\\ 
DeeBERT & 53.9 / 55.4 &77.2 & 67.6& 88.6& 68.5 \\ 
PABEE & 74.0 / 74.2&- &- &87.5 & -\\ 
CascadeRoBERTa & 78.9 / 78.1  & 86.8  & 70.5  & 90.8 & 81.0  \\ 
 ~~ + Vanilla KD  & 79.7 / 78.8  & \textbf{86.9} &  \textbf{70.8} & \textbf{91.4} &  \textbf{81.5} \\
% Possible add KD here to further enhance the performance 
\bottomrule
\end{tabular}
\caption{Test results from the GLUE server with RoBERTa models in our cascade framework. The speed-up ratio is approximately $3\times$ ($\pm4$\%). 
The - denotes unavailable results of PABEE.
}
\label{tab:roberta}

\end{table}
\section{Related Work}
% Our work aims to accelerate large-scale pre-trained language models inference while maintaining the superior performance.
% Previous efforts for accelerating the inference of PLMs can be mainly categorized to:

\medskip
\noindent\textbf{Model-level compression}
% 压缩和我们的方法是正交的 他们的技术可以拿来提升我们的性能
% aims to obtain a computation-efficient model, 
includes knowledge distillation (KD), pruning and quantization. 
KD focuses on transferring the knowledge from a large teacher model to a compact student model~\citep{Hinton2015Distilling}.
% , and various KD techniques have been applied to pre-trained language models for a more compact student model.
\citet{Sanh2019DistilBERT} propose DistilBERT and
\citet{Sun2019PatientKD} enhance KD by aligning the internal representations of the student and the teacher model. 
% A more comprehensive alignment strategy of layer mappings is proposed by \citet{li2020bertemd}.
Besides, \citet{Jiao2019TinyBERT} propose TinyBERT via a two-stage KD on augmented data.
Pruning methods deactivate the unimportant structures in the model like attention heads~\citep{voita2019analyzing,michel2019sixteen} and layers~\citep{fan2019layerdrop}.
Quantization methods target at using fewer physical bits to efficiently represent the model~\citep{zafrir2019q8bert,Shen2020QBERT,zhang2020ternarybert}.
We do not compare pruning and quantization methods since these techniques are orthogonal to our framework.
% with can be incorporated into to further boost the performance. 

\medskip
\noindent\textbf{Instance-level speed-up} accelerates the inference via adapting the computation according to the instance complexity~\citep{graves2016adaptive}. 
A representative framework is dynamic early exiting, which has been verified in natural language understanding~\citep{Xin2020DeeBERT,schwartz-etal-2020-right,Liu2020FastBERT, zhou2020pabee, liao2021global,sun2021early}, sequence labeling~\citep{li2021accelerating}, question answering~\citep{soldaini-moschitti-2020-cascade} and document ranking~\citep{xin-etal-2020-early}.
In this paper, we probes the work mechanism of dynamic early exiting, and find that it faces a serious performance bottleneck under high speed-up ratios. 
To remedy this, we generalize the idea to a model cascade and prove it is effectiveness even under high speed-up ratios for various natural language understanding tasks.
Concurrently with our work, \citet{Enomoto2021LearningCascade} adopt the similar idea and achieve better inference efficiency on image classification tasks.
% , demonstrating the generalizability of our framework.

% \medskip
% \noindent\textbf{Cascading Framework}

\section{Conclusion} % and Future Work}
In this paper, we point out that current dynamic early exiting framework faces a performance bottleneck under high speed-up ratios, due to insufficient shallow layer representations and poor exiting decisions of the internal classifiers.
To remedy this, we propose CascadeBERT, a model cascade framework with difficulty-aware regularization for accelerating the inference of PLMs. 
Experimental results demonstrate that our proposal achieves substantial improvements over previous dynamic exiting methods.
Further analysis validates that the framework is generalizable and produces more calibrated results.
% In the future, we hope to apply our method to accelerate generative models like GPT and BART.

\nocite{radford2019gpt,lewis2020bart}
% Extensive analysis further demonstrates the effectiveness and generalizability of our method.
% The difficulty-aware objective can further calibrate the model and thus make the exiting decisions more reliable.
% Further analysis shows that the proposed difficulty-awared regularization can help acceleration and calibrate model predictions. 
% explore more backbone architectures for evaluating the generalizability of our framework and
% In the future, we hope to apply the cascading framework for accelerating large-scale generative models like GPT.
% a finer-grained architecture search based on instance complexity in the future.
%, e.g., dynamic transformer layer module and heads selection, can be explored for future work.

\section*{Acknowledgements}
We thank all the anonymous reviewers for their constructive comments, Xuancheng Ren and Hua Zheng for their valuable suggestions in preparing the manuscript, and Wenkai Yang for providing the theoretical analysis.
This work was supported by a Tencent Research Grant. 
Xu Sun is the corresponding author of this paper.

\bibliographystyle{acl_natbib}
\bibliography{emnlp2021}

\appendix
\section{Comparison with FastBERT}
\label{apx:fastbert}
Performance comparison of our CascadeBERT with FastBERT is shown in Table~\ref{tab:fastbert} under $3.00\times$ and $4.00\times$ speed-up ratios.
Note that FastBERT utilizes a complicate internal classifier with self-attention mechanism and take the hidden states of all the sequence tokens for making predictions, while we only adopt the original linear-based classifier.
Our method can still outperform the FastBERT under high speed-up ratios, especially on tasks that require high-level semantic reasoning ability like MNLI.
\begin{table}[ht!]
\centering
\scalebox{0.7}{
\setlength{\tabcolsep}{3.5pt}
\begin{tabular}{@{}l@{\hspace{2pt}}l|c|c|c|c|c@{}}
\toprule
\multicolumn{2}{c|}{\textbf{Method}} & \textbf{MNLI-m} / \textbf{-mm}  &  \textbf{QNLI} & \textbf{QQP} & \textbf{SST-2} & \textbf{Average} \\
% \midrule 
% \multirow{2}{*}{\rotatebox[origin=c]{90}{$\sim$2$\times$}}  & CascadeBERT & 83.0 / 81.6 & 89.4  &  71.2 & 91.7 &  83.4 \\ 
% & ~~ -  w/o DAR & 82.0 / 81.7 & 88.8 &  71.0& 91.9&  83.1\\ 
\midrule 
\multirow{2}{*}{\rotatebox[origin=c]{90}{$\sim$3$\times$}}&FastBERT & 79.8  /  78.9 &88.2  & 71.5 &  92.1& 82.1  \\ 
& CascadeBERT &  81.2 / 79.5  & 88.5 & 71.0 & 90.9 & 82.2 \\ 
\midrule 
\multirow{2}{*}{\rotatebox[origin=c]{90}{$\sim$4$\times$}}&FastBERT &  76.1 / 75.2  & 86.4 & 70.5 &90.7 & 79.8 \\ 
& CascadeBERT &  79.3 / 77.9  & 86.5 & 70.0 & 90.3 & 80.8 \\ 
\bottomrule
\end{tabular}}
\caption{Performance comparison with FastBERT.
}\label{tab:fastbert}

\end{table}

\section{Expected Calibration Error}
\label{apx:ece}
Calibration measures the consistency between predictions' confidence and accuracy. 
A well calibrated model can be more reliable, e.g., it can give us a hint that it knows what it does not know, and thus it is easier for deployments in real-world applications.
It is formally expressed as a joint distribution $P(Q, Y)$ over confidences $Q \in R$ and labels $Y \in L$. 
When $P(Y = y \mid Q = q) = q $, the model is perfectly calibrated. For example, if the average confidence score of $100$ instances is $0.8$, there should be $80$ instances that are correctly predicted.
This probability can be approximated by grouping predictions into $k$ disjoint and equally-sized bins, where each bin consists of $b_k$ predictions.
The expected calibration error is defined as a weighted average of difference between each bin's accuracy ($\text{acc}(\cdot)$) and prediction confidence ($\text{conf}(\cdot)$):
\begin{equation}
    \text{ECE} = \sum_k \frac{b_k}{n} | {\rm acc}(k) - {\rm conf} (k)| 
\end{equation}
where $n$ is the number of total instances. 
A lower ECE denotes the model is better calibrated.
In this paper, we set $k=10$ for calculating the ECE score.

\section{Analysis for More Models in Cascade}
\label{apx:theory_more_teacher}
% 推导的定义
Suppose there are $n$ models $\{M_{1}, \ldots,M_{n}\}$ sorted from the smallest to largest according to number of layers in our cascade, with corresponding number of layers $\{L_{1}, \ldots, L_{n} \}$ and the task performance, e.g., classification accuracy $\{a_{1},\cdots,a_{n}\}$,
we want to explore whether incorporating another complete model into the original cascade can further improve the task performance and speed-up trade-off. In more detail, we propose to evaluate the difference of classification accuracy between the original cascade and the new cascade, under the same speed-up ratio. 
We denote the new added model as $M^{*}$ with classification accuracy $a^{*}$ consisting of $L^{*}$ layers, $L_{i} < L^{*} < L_{i+1}$ for a specific $i$.
Considering the instance emitting distribution, we denote the number of instances exiting after model $M_{j}$ ($j=1,\ldots,n$) as $s_{j}$ in the original $n$ models cascade. For the new $n+1$ models cascade, the number of samples exiting after model $M_{j}$ ($j=1,\ldots,n$) is $\hat{s}_{j}$ and there will be $\hat{s}^{*}$ instances emitting from $M^{*}$. 
Besides, we assume that the accuracy $a_{i}$ of $M_{i}$ is the same for any subsets of the original dataset.
The performance difference thus can be written as:
\begin{equation}
\label{ori target}
    T= \frac{1}{N}\left( \sum\limits_{k=1}^{n}a_{k} \hat{s}_{k} + a^{*}\hat{s}^{*} - 
    \sum\limits_{k=1}^{n}a_{k}s_{k} \right)
\end{equation}
under the conditions of 
\begin{equation}
\small
\label{ori_constraint}
\begin{cases}
%0& \text{x=0}\\
%1& \text{x!=0}
\sum\limits_{k=1}^{n}s_{k}  = \sum\limits_{k=1}^{n}\hat{s}^{k} + \hat{s}^{*} =N \\
\begin{aligned}
\sum\limits_{k=1}^{n}s_{k}L^{k}  &=  \sum\limits_{k=1}^{i} \hat{s}_{k} L^{k} +\hat{s}^{*}(L^{i}+L^{*})  + \\ &\quad \sum\limits_{k=i+1}^{n}\hat{s}_{k}(L^{k} + L^{*}) 
\end{aligned}\\
\end{cases} 
\end{equation}
where $L^{k} = \sum\limits_{i=1}^{k}L_{i}$ is the actual layer cost with the computation overhead and $N$ is the number of test instances.
The first condition indicates the total number of test instances is the same, and the second one guarantees that the total layer cost is same thus the speed-up ratio is identical.

% We need an assumption here that the accuracy $a_{i}$ of $M_{i}$ is the same for any subset of the original dataset.

There are infinite solutions for the above system of equations, as we can adjust the exiting thresholds of different models to achieve the same speed-up ratio.
% As we can expect, it is troublesome if we change all model's thresholds. 
We propose to simplify this by making a assumption that we only adjust the thresholds of $M_{i}$, $M_{i+1}$ and $M^{*}$ to achieve the same speed-up ratio, thus the following equation holds:
\begin{equation}
\label{assump2}
\hat{s}_{k} = s_{k},\quad k=1,2,\cdots, i-1, i+2, \cdots, n
\end{equation} 
Conditions in Eq.~(\ref{ori_constraint}) can thus be re-written as
\begin{equation}
\label{new_constraint}
\begin{cases}
%0& \text{x=0}\\
%1& \text{x!=0}
s_{i} + s_{i+1}  = \hat{s}_{i} + \hat{s}_{i+1}+\hat{s}^{*} \\
\begin{aligned}
s_{i}L^{i} + s_{i+1}L^{i+1} &=  \hat{s}_{i} L^{i} +\hat{s}^{*}(L^{i}+L^{*}) \\
&\qquad + \hat{s}_{i+1}(L^{i+1} + L^{*}) \\
\end{aligned}\\
\end{cases} 
\end{equation}
Then we can further calculate $s_{i}$ and $s_{i+1}$ as 
\begin{equation}
\label{s_i s_i+1}
\begin{cases}
%0& \text{x=0}\\
%1& \text{x!=0}
s_{i}   = \hat{s}_{i} +\hat{s}^{*} -\frac{L^{*}}{L_{i+1}}(\hat{s}_{i+1} +\hat{s}^{*}) \\
s_{i+1} = \hat{s}_{i+1} + \frac{L^{*}}{L_{i+1}}(\hat{s}_{i+1} +\hat{s}^{*}) 
\end{cases} 
\end{equation}
By plugging the equations in Eq.~\ref{s_i s_i+1} into the Eq.~\ref{ori target}, and use the assumption in Eq.~\ref{assump2} we get
{
\resizebox{\linewidth}{!}{\parbox{\linewidth}{\begin{align*}
\label{new target}
T &= \frac{1}{N}\left[ a_{i}(\hat{s}_{i} - s_{i}) +  a_{i+1}(\hat{s}_{i+1} -s_{i+1})+ a^{*} \hat{s}^{*} 
%- (a_{i}s_{i} +  a_{i+1}s_{i+1})
\right] \\ & =
\frac{1}{N} \left[ 
a^{*} \hat{s}^{*} +a_{i}(\frac{L^{*}}{L_{i+1}}(\hat{s}_{i+1} + \hat{s}^{*}) - \hat{s}^{*} )\right. \\
&\left. \qquad\qquad -a_{i+1} \frac{L^{*}}{L_{i+1}}(\hat{s}_{i+1} +\hat{s}^{*}) 
\right] \\ 
&= \frac{1}{N} \left[ 
\hat{s}^{*}(a^{*} - a_{i}) - \frac{L^{*}}{L_{i+1}}(\hat{s}_{i+1} +\hat{s}^{*})(a_{i+1}-a_{i})  
\right]
\end{align*}}}
}%
The final expected performance difference is thus:
\begin{equation}
\begin{aligned}
T(\hat{s}^{*}, L^{*}) &= \frac{1}{N} \biggl[ 
\hat{s}^{*}(a^{*} - a_{i}) - \\
&  \qquad  \frac{L^{*}}{L_{i+1}}(\hat{s}_{i+1} +\hat{s}^{*})(a_{i+1}-a_{i}) \biggr]
\label{final_target}
\end{aligned}
\end{equation}
where the index $i$ satisfies that $L_{i} < L^{*} < L_{i+1}$.
%Notice that the variable $a^{*}$ depends on $l^{*}$,
Note that the model accuracy $a^{*}$ is related to the size $L^{*}$ of model, as a larger model with more layers tends to achieve a better task performance. 
It indicates that the performance difference depends on the number of samples exits at model $M^{*}$ ($\hat{s}^{*}$), and the layers of $M^{*}$ ($L^{*}$). 
%and the classification accuracy of model $M^{*}$ ($a^{*}$).
% $\hat{s}^{*}$ depends on the adjustment of the exiting thresholds of model $M_{i}$, $M_{i+1}$ and new model $M^{*}$, $l^{*}$ decides the the classification accuracy $a^{*}$.
% Since we can choose different $l^{*}$ satisfies $l_{i} < l^{*} < l_{i+1}$ for different $i$ , we want to explore the maximum performance gain when we add a new model under the condition of already have $n$ models. That is, we now focus on the following target:
% \begin{equation}
% \small
% \label{max_gain}
% T_{max} = \max\limits_{i}\max\limits_{\hat{s}^{*}, l^{*}}\left\{ T(\hat{s}^{*}, l^{*}) \big|  l_{i} < l^{*} < l_{i+1} \right\}
% \end{equation}
If we fix the index $i$ when we add the new model $M^{*}$, since we have $a_{i} \leq a^{*} \leq a_{i+1}$, from Eq~(\ref{final_target}) we can get 
% {\small
\begin{align*}
& T(\hat{s}^{*}, L^{*}) \\ & \leq \frac{1}{N} \left[ 
\hat{s}^{*}(a^{*} - a_{i}) - \frac{L^{*}}{L_{i+1}}(\hat{s}_{i+1} +\hat{s}^{*})(a^{*}-a_{i})  
\right]  \\ & \leq
\frac{1}{N}(a^{*} - a_{i})\left[ \hat{s}^{*} -
\frac{L^{*}}{L_{i+1}}(\hat{s}_{i+1} +\hat{s}^{*})
\right] 
% \\ & \leq
% \frac{1}{N}(a^{*} - a_{i}) \hat{s}^{*} \left( 1 -
% \frac{l^{*}}{l_{i+1}}
% \right) \\ & \leq
% \frac{\hat{s}^{*}}{N}(a^{*} - a_{i}) \left( 1 -
% \frac{l^{*}}{l_{i+1}}
% \right) 
% \\ & \leq
% \frac{s_{i} + s_{i+1}}{N}(a^{*} - a_{i}) \left( 1 -
% \frac{l^{*}}{l_{i+1}}
% \right)
\end{align*}
% }
On the one hand, as $L^{*} \rightarrow L_{i+1}$, $(a^{*} - a_{i})$ will increase to $(a^{i+1} - a_{i})$, but $\hat{s}^{*} -
\frac{L^{*}}{L_{i+1}}(\hat{s}_{i+1} +\hat{s}^{*})$ will decrease to $-\hat{s}_{i+1}$; On the other hand, when $L^{*}$ gets close to  $L^{i}$, $a^{*} - a^{i} \rightarrow 0$. 
This trade-off indicates that the layer size of $M^{*}$ should be carefully chosen to achieve performance improvements. 
Otherwise, the overall gain could be negative. Besides, the upper bound of maximum gain also depends on the number of samples exit at $M^{*}$. 
Thus, adjusting thresholds properly is also important.
Additionally, we can further scale the upper bound as:
% {\small
\begin{align*}
& T(\hat{s}^{*}, L^{*})  \\ & \leq
\frac{1}{N}(a_{i+1} - a_{i})\left[ \hat{s}^{*} -
\frac{L_{i}}{L_{i+1}}(\hat{s}_{i+1} +\hat{s}^{*})
\right]
\\ & \leq
\frac{s_{i} + s_{i+1}}{N}(a_{i+1} - a_{i})\left( 1 -
\frac{L_{i}}{L_{i+1}}
\right)
\end{align*}
% }
which indicates that
{\small
\begin{align*}
& \max\limits_{s^{*}, L^{*}}\left\{T(\hat{s}^{*}, L^{*}) \right\} 
\\ & \leq
\frac{s_{i} + s_{i+1}}{N} \left(\max\limits_{i}\left\{a_{i+1} - a_{i} \right\}\right)
\left( 1 -
\min\limits_{i}\left\{\frac{L_{i}}{L_{i+1}} \right\}
\right).
\end{align*}
}
Note that 
\begin{equation}
  \max\limits_{i}\left\{ a_{i+1} - a_{i} \mid  M_{i},\cdots, M_{n} \right\}  
\end{equation}
and
\begin{equation}
\min\limits_{i}\left\{  \frac{L_{i}}{L_{i+1}} \mid M_{i},\cdots, M_{n} \right\}
\end{equation}
are non-increasing as $n$ gets larger. It means the maximum expected performance gain of adding another model can be marginal as the number of models in the original cascade becomes larger.

In all, our analysis shows that we should carefully select model $M^{*}$ with layers $L^{*}$, and tune the exiting threshold to adjust number of samples exit after $M^{*}$, to guarantee that the target in Eq.~\ref{final_target} is positive, in order to gain improvements by incorporating more models into the original cascade. 

\end{document}